**Main Manuscript for**

# Invariant neuromorphic representations of tactile stimuli improve robustness of a real-time texture classification system

Mark M. Iskarous[a,b,*], Zan Chaudhry[a,c], Fangjie Li[a,d], Samuel Bello[a], Sriramana Sankar[a], Ariel Slepyan[e], Natasha Chugh[f,g], Christopher L. Hunt[a,h], Rebecca J. Greene[e], Nitish V. Thakor[a,e]

[a] Department of Biomedical Engineering, Johns Hopkins School of Medicine, Baltimore, MD, 21205, USA.
[b] Department of Organismal Biology and Anatomy, University of Chicago, Chicago, IL, 60637, USA.
[c] National Heart Lung and Blood Institute, National Institutes of Health, Bethesda, MD, 20814, USA.
[d] Department of Biomedical Engineering, Vanderbilt University, Nashville, TN, 37325, USA.
[e] Department of Electrical and Computer Engineering, Johns Hopkins University, Baltimore, MD, 21218, USA.
[f] Department of Public Health, Johns Hopkins University, Baltimore, MD, 21218, USA.
[g] Texas College of Osteopathic Medicine, The University of North Texas Health Science Center, Fort Worth, TX, 76107, USA.
[h] Infinite Biomedical Technologies, LLC., Baltimore, MD, 21202, USA.
* Mark M. Iskarous, 5496 S Hyde Park Blvd, Apt 701, Chicago, IL, 60615, USA, (408) 960-5866, miskarous@uchicago.edu

All work was done at Johns Hopkins University. All other affiliations are changes since the work was completed.

**Author Contributions:**

Conceptualization: M.M.I., N.V.T.
Methodology: M.M.I., Z.C., F.L., S.B., S.S., A.S., C.L.H., R.J.G.
Software: M.M.I., Z.C., F.L., S.B.
Validation: M.M.I., Z.C., F.L., S.B.
Formal Analysis: M.M.I., Z.C., F.L. S.B.
Investigation: M.M.I., Z.C., F.L., S.B.
Resources: M.M.I., Z.C., F.L., S.B., S.S., A.S.
Data Curation: M.M.I., Z.C., F.L., S.B.
Writing – original draft: M.M.I.
Writing – review and editing: M.M.I., N.V.T.
Visualization: M.M.I., Z.C, S.B., S.S., N.C.
Supervision: N.V.T.
Project Administration: M.M.I., N.V.T.
Funding Acquisition: M.M.I, N.V.T.

**This PDF file includes:**

    Main Text
    Figures 1 to 4
    Table 1

**Abstract**

Humans have an exquisite sense of touch which robotic and prosthetic systems aim to recreate. We developed algorithms to create neuron-like (neuromorphic) spiking representations of texture that are invariant to the scanning speed and contact force applied in the sensing process. The spiking representations are based on mimicking activity from mechanoreceptors in human skin and further processing up to the brain. The neuromorphic encoding process transforms analog sensor readings into speed and force invariant spiking representations in three sequential stages: the force invariance module (in the analog domain), the spiking activity encoding module (transforms from analog to spiking domain), and the speed invariance module (in the spiking domain). The algorithms were tested on a tactile texture dataset collected in 15 speed-force conditions. An offline texture classification system built on the invariant representations has higher classification accuracy, improved computational efficiency, and increased capability to identify textures explored in novel speed-force conditions. The speed invariance algorithm was adapted to a real-time human-operated texture classification system. Similarly, the invariant representations improved classification accuracy, computational efficiency, and capability to identify textures explored in novel conditions. The invariant representation is even more crucial in this context due to human imprecision which seems to the classification system as a novel condition. These results demonstrate that invariant neuromorphic representations enable better performing neurorobotic tactile sensing systems. Furthermore, because the neuromorphic representations are based on biological processing, this work can be used in the future as the basis for naturalistic sensory feedback for upper limb amputees.

**Significance Statement**

We developed algorithms to create neuron-like (neuromorphic) spiking representations of texture that are invariant to the scanning speed and contact force applied in the sensing process. The algorithms were tested on a tactile texture dataset and in a real-time human-operated texture classification system. Texture classification systems built on the invariant representations have higher classification accuracy, improved computational efficiency, and increased capability to identify textures explored in novel conditions. These results demonstrate that invariant neuromorphic representations enable better performing neurorobotic tactile sensing systems. Furthermore, because the neuromorphic representations are based on biological processing, this work can be used in the future as the basis for naturalistic sensory feedback for upper limb amputees.

**Main Text**

**Introduction**

Upper limb amputees have expressed the importance of sensory feedback in maximizing the functional capabilities of their neuroprostheses (1). In particular, tactile sensing is critical for the exploration and manipulation of the user's environment (2). Therefore, engineers have aspired to create artificial sensing systems that match the capabilities of human tactile sensing (3) including the perception of material properties such as roughness, compliance, coldness, and friction (4). The strategy for providing this sensory information to the amputees consists of developing electronic skins (e-skins) that incorporate tactile sensors (5–8) and then interfacing with the nervous system to restore naturalistic tactile sensations (9–16).

In order to appropriately interface with the nervous system, an understanding of the biological encoding of tactile information is required. Broadly speaking, the sensation of touch is mediated by two categories of mechanoreceptors in skin: slowly adapting (SA) mechanoreceptors (composed of Merkel cells and Ruffini endings) and rapidly adapting (RA) mechanoreceptors (composed of Meissner and Pacinian corpuscles) (17). The SA mechanoreceptors primarily produce spiking activity corresponding to the strength of sustained tactile events, whereas RA mechanoreceptors produce spiking activity corresponding to large changes in tactile events (such as onset/offset or vibration). These mechanoreceptors are further categorized as Type I which are shallower in the skin and have smaller receptive fields and Type II which are deeper and have larger receptive fields (3). Therefore, most mechanoreceptors in skin are categorized as SAI, SAII, RAI, or RAII. The activity produced by the population of mechanoreceptors encodes information about tactile stimuli such as form and texture (18), edge orientation (19), and fingertip force direction (20).

In this paper, we focus on the encoding of texture stimuli because of their rich spatiotemporal structure. In particular, we investigate how the encoding of texture changes along two dimensions: the scanning speed with which a texture is explored, and the amount of contact force applied when exploring a texture. There is perceptual invariance for both scanning speed and contact force (21, 22) meaning that the texture is perceived as the same texture despite differences in how it was explored. Perceptual invariance is based on invariance in neural representation which can be achieved at different points along the tactile processing chain. The neural representation of texture in the somatosensory cortex is invariant to scanning speed (23) even though the activity at the nerve fibers contracts or dilates in time as the scanning speed increases or decreases, respectively (24). This implies that speed-invariant texture representation is achieved in the central nervous. In contrast, the activity of nerve fibers in response to changes in contact force is relatively unaffected (25) which implies that force-invariant texture representation is achieved locally at the receptors.

Previous work in artificial texture sensing has focused on the classification of different textures directly using tactile sensor readings with both supervised (26) and unsupervised learning methods (27). Other work has first transformed the analog sensor readings into neuron-like patterns of spiking activity (which we define as *neuromorphic*) and then classified those texture patterns (28–35). In other touch-related tasks such as edge detection and braille letter reading, neuromorphic software and hardware encoding techniques have been incorporated into robotic systems (neurorobots) to improve their performance and computational efficiency (36–42). Neuromorphic systems have additionally been used for sensory neuroprosthesis applications (43, 44) such as human texture discrimination through electrical stimulation of nerves (45, 46).

We have previously built a flexible multilayer tactile sensor (47). We transformed texture data from the sensor into neuromorphic activity that mimics the behavior of SA and RA mechanoreceptors using the Izhikevich neuron model (48) and classified textures using an unsupervised training algorithm (49). We did preliminary design and testing on a rotating drum apparatus (50) which we used in this work to collect a dataset of various textures explored at different scanning speeds and contact forces.

In this work, we incorporate novel bio-inspired algorithms in the neuromorphic encoding process that account for the scanning speed and contact force applied when exploring a texture to create speed and force invariant texture representations. These representations improve accuracy and computational efficiency of texture classification across texture exploration conditions. Beyond this, the invariant representations also enable the successful identification of texture in conditions that were not previously experienced. Finally, we integrate the speed invariance algorithm into a real-time human-operated texture classification system. Ultimately, this work demonstrates the value of neuromorphic texture representations in a neurorobotic application and paves the way for improved sensory feedback in an upper limb prosthesis system (Fig. 1A).

**Results**

The results are broken into four sections. The first section describes the completed rotating drum apparatus built to collect the tactile texture dataset used throughout the paper. The second section describes the algorithms developed to create the neuromorphic speed and force invariant texture representations. The signal processing of the tactile readings consists of three main processing stages which take inspiration from biological encoding: a force invariance module, a spiking activity encoding module, and a speed invariance module (Fig. 1B). These modules are the main novel developments of this work, and the remainder of the paper serves to computationally validate the value of the algorithms and apply them in a neurorobotic application. The third section describes a series of texture classification analyses that demonstrate that the invariance modules improve classification efficiency, accuracy, and robustness to novel speed-force conditions. The final section demonstrates the integration of the speed invariance module into a real-time human-operated texture classification system.

**Rotating Drum Apparatus and Tactile Texture Dataset**

A rotating drum apparatus was designed to apply 3D-printed textures to a tactile sensor array at programmed scanning speeds and applied forces (Fig. 2A). The preliminary design, control scheme, and validation of the rotating drum can be found in (50).

In this work, a set of 16 textures was designed and 3D-printed (Fig. 2B). The set was composed of 3 texture groups (circular ridges, rectangular ridges, and waves) and a smooth control texture. Each texture group had 5 variations: base, double height, double space, half height, and half space. Height refers to the height of the textural features which affects the amplitude of the measured force. Spacing refers to the distance between the textural features which affects the period of the measured force.

The texture set was applied to a multilayer tactile sensor based on (47). The sensor has two layers of a 3x3 grid of tactile pixels (which we define as *taxels*) spaced at 5 mm intervals. The sensor uses a piezoresistive fabric to transform force into changes in voltage. The top layer represents Type I mechanoreceptors, while the bottom layer represents Type II mechanoreceptors.

The rotating drum applied each texture at 15 speed-force combinations. Scanning speed was tested from 40 mm/s to 120 mm/s (in 20 mm/s increments) based on biological scanning speed measurements (51). Applied force (measured through a load cell at the base of the drum) was tested at 250, 500, and 1000 g. Further details about the dataset can be found in Materials and Methods. Movie S1 shows the drum in operation.

**Speed and Force Invariant Texture Representation**

The neuromorphic encoding process transforms analog sensor readings into speed and force invariant spiking representations in three sequential stages: the force invariance module (in the analog domain), the spiking activity encoding module (transforms from analog to spiking domain), and the speed invariance module (in the spiking domain). The force invariance module and the spiking activity encoding module together create a force invariant spiking representation of texture to match biological recordings in nerve fibers. The spiking activity encoding module takes the readings from each taxel and models both SA mechanoreceptors that respond to static levels of force magnitude and RA mechanoreceptors that respond to acute changes in force. The addition of the speed invariance module creates a force and speed invariant spiking representation of texture to match biological recordings in the cortex. The stages of signal processing are depicted in Fig. 1B. The force invariance and speed invariance modules are individually toggled on or off in the analyses to test the effectiveness of the algorithms.

In order to create force invariant spiking patterns (25), the input current to the spiking activity encoding module is scaled to account for the applied force. An empirical approach was used to determine the force scaling factors (justification of this approach and explanation of the procedure are detailed in Materials and Methods). Since the goal is to create force invariant spiking patterns, the spike rate from the SA encoding of each taxel was used to make the objective function for an optimization problem. The target spike rate was set to the spike rate at the 500 g applied force (averaged across all trials). Therefore, scaling coefficients are determined for each texture for each taxel for each speed at 250 and 1000 g. More formally, we solve for the matrix $C \in R^{16 \times 18 \times 5 \times 3}$ (whose entries $C_{i,j,k}^T$ are scaling coefficients for applied texture *T*, taxel *i*, speed setting *j*, and force setting *k*) such that:

$$|\mathrm{SR}(\Phi_{SA}(C_{i,j,k}^T R_{i,j,k}^T)) - SR(\Phi_{SA}(R_{i,j,500}^T))| < \epsilon$$

(1)

$R_{i,j,k}^T$ is the time series data from the tactile sensor for applied texture *T*, taxel *i*, speed setting *j*, and force setting *k*. The function $\Phi_{SA}(R)$ encodes the sensor readings as a spike train of SA mechanoreceptors using the Izhikevich model (described in the next paragraph and in further detail in Materials and Methods). The function SR(ST) calculates the mean spike rate of an input spike train (ST) (i.e. the number of spikes divided by the total time). The calculation of the force scaling coefficients is only done one time for the whole dataset. These coefficients are applied to the scaling factor in the spiking activity encoding but are only used for the SA model since SA mechanoreceptors more directly respond to static force, whereas RA mechanoreceptors respond to the dynamics of force (which are much less affected by a scaling factor).

The spiking activity encoding module uses the Izhikevich neuron model (48) to transform analog tactile sensor readings into neuromorphic spiking activity (the model equations and parameters are described in Materials and Methods). Both SA and RA mechanoreceptor models apply the Tonic Spiking model (52) to the tactile sensor readings except that the RA model smooths and differentiates the readings first. The SA model therefore produces spikes at a rate proportional to the sensor reading, whereas the RA model produces spikes at a rate proportional to changes in the sensor reading. The dataset produced is a set of spike trains which consist of an ordered list of spike times. An example of one taxel reading encoded as SA and RA neuron outputs is shown in Fig. 2C (and an animated version can be seen in Movie S2). The block diagram with more details for the spiking activity encoding module is shown in Fig. 2D.

The speed invariance module uses a mechanistic approach to create speed invariant spiking patterns (24). The output spike times from the spiking activity encoding module are scaled inversely proportional to the scanning speed (with a reference speed set to 120 mm/s). This means that spiking activity from a 120 mm/s scanning speed trial would not be scaled in time, but activity from a 40 mm/s scanning speed sample would be contracted by a factor of 3. Since the spike trains are represented as an ordered list of spike times, the spike times only need to be multiplied by the appropriate scalar value. More formally, a speed invariant spike train $(IST_{i,j,k}^T)$ is calculated from $ST_{i,j,k}^T$ (for applied texture *T*, taxel *i*, speed setting *j*, and force setting *k):*

$$IST_{i,j,k}^T = \frac{j}{120} ST_{i,j,k}^T$$

(2)

These three modules together create a speed and force invariant representation of texture in the spiking domain based on biological principles. Going forward into the texture classification results,

"Original" data refers to the spike trains produced when only the spike activity encoding module is used, "Speed Scaled" data refers to the spike trains produced when both the spiking activity encoding and speed invariance module are used, "Force Scaled" data refers to the spike trains produced when both the force invariance and spiking activity encoding module are used, and "Speed and Force Scaled" data refers to the spike trains produced when all three modules are used.

**Texture Classification in Trained and Novel Conditions**

Texture classification tasks are used to evaluate the speed and force invariant representations. First, simple spiking features are extracted from the spike train dataset. A time series of spike rates (SR) and spike counts (SC) are calculated over non-overlapping sliding windows for SA-encoded and RA-encoded spike trains, respectively. To make texture classification computationally feasible, the feature data is dimensionally reduced using principal component analysis (PCA) (53). All classification is done using linear discriminant analysis (LDA) (54) on the set of principal components (PCs) calculated from the feature data. In the classification process, the number of PCs varied. Further details about feature extraction, classification statistics, and cross-validation technique are in Materials and Methods.

Three classification analyses completed over a range of PCs are depicted in Fig. 3. The datapoints of the line plots (n = 20) are the mean classification accuracies (μ), and vertical error bars represent the standard deviation (σ).

The first texture classification analysis tests classification of each individual texture independent of exploratory conditions. The LDA model is trained with different labels for each texture (textures A-P, Fig. 2B) and asked to predict one of the 16 textures labels on the test data. The results of this texture classification analysis are shown in Fig. 3A. All datasets eventually saturate at similar accuracy levels (≈99.5%). At lower PCs, the "Speed and Force Scaled" data consistently has a higher classification accuracy than the other datasets with "Speed Scaled" next, "Force Scaled" after that, and "Original" being worst. These results indicate that individually, speed scaling and force scaling improve the representation of texture, and that they combine to improve the texture representation even further. The speed and force invariance algorithms successfully created invariant texture representations for a texture set that was explicitly designed to confound with the force and speed conditions.

The second texture classification analysis tests classification of similar groups of textures independent of exploratory conditions. The LDA model is trained with different labels for each texture group (smooth, circular ridges, rectangular ridges, waves) and asked to predict one of those 4 texture group labels on the test data. The results of the texture group classification analysis are shown in Fig. 3B. This analysis shows the same trend for the different datasets as before except that the different datasets do not all converge to the same final accuracy. Overall, this analysis examines how well the basic features of each texture group (smooth, circular ridges, rectangular ridges, waves) can be extracted even with variations in feature size, feature frequency, scanning speed, and applied force. Texture group classification is a more difficult problem (even though random chance is higher), and therefore the overall accuracies are less than in Fig. 3A.

The last texture classification analysis tests classification of each individual texture when confronted with novel exploratory conditions. The LDA model is trained with different labels for each texture (textures A-P) and asked to predict one of the 16 texture labels on the test data. The difference from the first analysis is that the training set is a subset of the speed-force combinations. Only 3 speeds (40, 60 and 80 mm/s) and 2 forces (250 and 500 g) are used for training which corresponds to 6 of the 15 speed-force combinations. The test set comes from all 15 speed-force combinations. The results of the texture extrapolation classification analysis are shown in Fig. 3C-F. The results are broken out by test condition to highlight the contributions of the speed and force invariance algorithms. In Fig. 3C, the test set is the 2 speed-force combinations where neither the speed nor the force is trained. In this scenario, the "Speed and Force Scaled" data has better saturation classification accuracies than the "Force Scaled" or "Speed Scaled" data which are both much better than the "Original" data. As would be expected, the classifier takes a performance hit when seeing speed-force combinations that it was not trained on, but the speed and force invariance algorithms make the representations significantly more robust to novel conditions. Confusion matrices that break down the class-by-class results of Fig. 3C are in Figs. S1 – S4. In Fig. 3D, the test set is the 3 speed-force combinations where the speed is trained, but the force is untrained. In this scenario, the two force scaled sets have much higher saturated classification accuracies than the two without force scaling. In Fig. 3E, the test set is the 4 speed-force combinations where the speed is untrained, but force is trained. In this scenario, the two speed scaled sets have much higher saturated classification accuracies than the two without speed scaling. Finally, in Fig. 3F, the test set is the same 6 speed-force combinations used in the training set and therefore the classification accuracies are very high (and the trend of the first two analyses persists).

In Table 1, the mean accuracy and standard deviation for all the analyses are reported for 50 PCs which approximately represents the knee of the classification accuracy curves.

**Real-time Texture Classification with Speed Invariant Representation**

As proof of concept, a demonstration of the speed scaling algorithm was implemented in a real-time system to classify a set of 5 textures (Movie S3). Flat texture plates were 3D-printed corresponding to textures A (smooth), B (circular ridges), D (circular ridges with double space), L (waves), and N (waves with double space) in Fig. 2B. A human operator applied a 3x3 tactile sensor array attached to a 3D-printed finger across the textures with an infrared motion tracker attached to their wrist to calculate their scanning velocity. The operator's movement is unconstrained, but they purposefully varied the intended velocity profile (slow, medium, fast, slow-to-fast, fast-to-slow). The speed invariance module and feature extraction are done on a microcontroller and classification is done on a computer. For the demonstration, the predicted texture is shown on a computer screen at the end of the texture scan. The real-time texture classification system diagram is shown in Fig. 4A. Further details about the processing pipeline, training dataset, speed invariance module, feature extraction, testing protocol, and statistical methods are in Materials and Methods.

Three classification analyses completed over a range of PCs are depicted in Fig. 4B-C. The datapoints of the line plots (n = 300) are the mean classification accuracies ($\mu$), and vertical error bars represent the standard error ($\sigma_{\bar{x}}$).

The first texture classification analysis tests classification of each individual texture independent of exploratory conditions. This is equivalent to the analysis of Fig. 3A but specific to this dataset of 5 textures and 5 velocity profiles (and collected with the human-operated system). The results of the texture classification analysis of this offline dataset are shown in Fig. 4B (blue lines). The "Speed Scaled" data had a significantly higher accuracy than the "Original" data at all PCs ($p < 0.01$ and $d > 1$). The "Speed Scaled" data saturated at around 90% classification accuracy while the "Original" data saturated at around 65%.

The second analysis explores how a classifier trained on a dataset from a prior session will respond to data from a new session. The results are shown in Fig. 4B (green lines). Both the "Speed Scaled" and "Original" data converge to a similar saturation accuracy around 50%, but the "Speed Scaled" data can reach the plateau much faster (around 8 PCs) and therefore has a significantly higher accuracy ($p < 0.01$ and $d > 1$) for fewer PCs (from 5 to 30). Additionally, the peak accuracy of the "Original" data is 50.45% (at 54 PCs), which the "Speed Scaled" data exceeds at lower PCs (10 – 26).

As with the offline dataset, we wanted to see how robust the real-time classifier would be when confronted with novel exploratory conditions. The LDA model is now only trained on a subset of the velocity profiles (slow, medium) but tested with all of them. The results of the texture extrapolation classification analysis are shown in Fig. 4C. The blue lines show the tested set consisting of the velocity profiles used to train the classifier. The peak accuracy of the "Original" data is 38.6% (at 20 PCs), which the "Speed Scaled" data exceeds starting from 5 PCs (and has a peak accuracy of 48.1% at 18 PCs). The "Speed Scaled" data has significantly higher accuracy ($p < 0.01$ and $d > 1$) at lower PCs (4 – 16) and has significance but only a "medium" effect size at the others ($p < 0.01$ and $0.5 < d < 1$). The green lines show the tested set consisting of the velocity profiles not used to train the classifier. The peak accuracy of the "Original" data is 28.2% (at 23 PCs), which the "Speed Scaled" data exceeds starting from 4 PCs (and has a peak accuracy of 42.7% at 9 PCs). The "Speed Scaled" data has significantly higher accuracy ($p < 0.01$ and $d > 1$) at most PCs (4 – 22) and has significance but only a "medium" effect size at the others ($p < 0.01$ and $0.5 < d < 1$).

Finally, the trained LDA model was deployed in a real-time system to classify incoming texture data (demonstrated in Movie S3). To maximize performance, the LDA model was trained with the entire training dataset. This contrasts with the analyses of Fig. 4B-C which randomly sampled the training dataset for the purpose of cross-validation and system characterization. The real-time classification was done using 25 PCs, but the dataset was used to simulate the system operating for all other number of PCs. The results are shown in Fig. S5. Both the "Speed Scaled" and "Original" data converge to a similar saturation accuracy around 52%. The peak accuracy of the "Original" data is 52.33% (at 67 PCs), which the "Speed Scaled" data exceeds for most PCs starting from 7 (and has a peak accuracy of 61.67% at 25 PCs). The "Speed Scaled" data has significantly higher accuracy ($p < 0.01$ and $h > 0.5$) at many of the lower PCs (6 – 26). Additionally, the Speed Scaled" data has significantly higher accuracy for most analyses using 34 PCs or fewer ($p < 0.05$ and $0.2 < h < 0.5$).

**Discussion**

Inspired by biological tactile processing (Fig. 1A), this work developed scanning speed and contact force invariance algorithms to create invariant spiking representations of texture stimuli (Fig. 1B). A tactile texture dataset was collected with a rotating drum apparatus (Fig. 2A-B, Movie S1) that enabled precise control of scanning speed and contact force when interacting with a variety of 3D-printed textures. The rotating drum is a useful tool that could also be leveraged to characterize tactile sensor arrays or further explore texture sensing systems. The development of neuromorphic spiking representations of texture (Figs. 1B, 2C-D, and Movie S2) serves two purposes. Firstly, as demonstrated in this work, mimicking biological sensory processing can illuminate new pathways to improve artificial sensing systems (in this case, texture classification). Secondly, mimicking biological sensory processing allows for more seamless integration between artificial and biological systems such as a neuroprosthesis that provide sensory feedback to upper limb amputees (Fig. 1A). The value of these representations was validated first with the tactile texture dataset and then deployed in a real-time human-operated texture classification system.

The classification analyses of the offline tactile texture dataset (Figs. 3, S1-S4, and Table 1) showed the value of the speed and force invariant representations. First, classification accuracy of individual textures (in any exploratory condition) increased with the inclusion of either invariance module and increased further with both invariance modules included (Fig. 3A). This implies that the invariant representations are better at extracting characteristics of the textures by reversing the effect of scanning speed and contact force on the sensor readings. Second, classification accuracy of related textures (in any exploratory condition) similarly improved with the inclusion of the invariance modules (Fig. 3B). This further implies that the invariant representations are better able to find the essence of texture features (smooth, circular ridges, rectangular ridges, waves) even when the features have different sizes and spacing. Third, classification accuracy of individual textures when confronted with novel exploratory conditions was improved with the inclusion of the invariance modules (Fig. 3C-F). Specifically, the force invariance module improved classification accuracy when confronted with novel forces (Fig. 3C-D), and the speed invariance module improved classification accuracy when confronted with novel speeds (Fig. 3C and 3E). This is a critical result for application of these algorithms in real-world environments because it means that a texture classification system does not need to be trained for all possible situations it might encounter. This greatly simplifies the development process and improves the utility of an artificial texture sensing system. Additionally, it enables the extension of the system to human-operated exploration which is less precise and has dynamically changing scanning speed and contact forces. Finally, across all these results, the invariance modules can achieve similar classification accuracies (as systems without the invariance modules) using a smaller number of PCs. This means that the invariance modules can improve the computational efficiency of texture classification.

A real-time human-operated texture classification system was developed to demonstrate the real-world applicability of the findings from the tactile texture dataset (Figs. 4, S5, and Movie S3). The system used a motion tracker to calculate scanning speed of an unconstrained operator moving the finger sensor with different velocity profiles. A microcontroller converted the tactile sensor readings to neuromorphic spiking activity, applied the speed invariance algorithm, and extracted spiking features that were then classified on a computer (Fig. 4A). The results (Fig. 4B-C) showed that the speed invariance module significantly improved classification accuracy while using fewer PCs, and improved classification robustness to new exploratory conditions (consistent with the offline analyses). These classification accuracies were lower than those for the tactile texture dataset because dynamic velocities introduce more error in the system and because session-to-session variability in the sensor readings limit the similarity between training and testing data. Indeed, the results highlight the fundamental difference between a human-operated real-time system and an offline dataset collected with a robot. It would be expected that the relationship between the blue lines in Fig. 4B and C would match those in Fig. 3A and F (i.e. that at high PCs the "Original" and "Speed Scaled" data saturate to a similar accuracy) because the training and test set come from the same stimuli (combination of texture and velocity profile). In reality, the blue lines look more like those in Fig. 3C-E with the "Original" and "Speed Scaled" data saturating at different accuracy levels because the speed scaling helps the classifier when confronted with novel exploratory conditions. This demonstrates that to the classifier, the variation in velocity profiles (and potentially other factors such as the path of texture scanning) due to human imprecision is seen as a novel condition. Therefore, the invariance algorithm is even more critical to help the classifier operate in real-world situations. As a final note of applicability, by tracking motion with an embedded inertial measurement unit and moving the classification algorithm to the microcontroller, this system can be adapted to an embedded or mobile application.

A limitation of this work is that the force invariance module was difficult to adapt to a real-time system. There are a couple reasons for this. Firstly, there was no straightforward way to get a ground truth measurement of force in real-time that could be used in a force invariance algorithm (for speed invariance ground truth was provided by the motion tracker). Secondly, because an empirical approach was taken to determine the appropriate force scaling coefficients for an offline dataset, the force scaling coefficients are ineffective if anything about the tactile sensor changes (which naturally occurs every time data is collected). To make a real-time force invariance module possible a high-precision tactile sensor array is needed whose characteristics are stable over time.

Ultimately, this work demonstrated that biologically inspired invariant representations can enable better performing neurorobotic tactile sensing systems that can be used for a wide range of applications. Furthermore, because the texture representations are based on the spiking patterns of SA and RA mechanoreceptors in skin, this work also has applicability in the domain of sensory feedback for upper limb amputees. The texture representations more closely resemble biological spike activity and therefore stimulation patterns based on these representations can produce more naturalistic tactile sensations. As with this neurorobotic application, the amputee can explore tactile stimuli in a more natural and uncontrolled manner while still receiving consistent stimulation patterns to discriminate between stimuli more reliably.

## Materials and Methods

### Experimental Design
This work consists of two experiments. The first experiment uses a tactile texture dataset collected with a rotating drum apparatus we built to validate the benefits of neuromorphic speed and force invariant texture representations. A set of 16 textures is applied to a 3x3x2 tactile sensor array with 15 speed-force combinations. The textures are classified to show that the invariant representations improve classification efficiency, accuracy, and robustness to novel exploratory conditions. In the second experiment, the speed invariance algorithm is integrated into a real-time human-operated texture classification system. The speed invariance algorithm similarly improves texture classification accuracy in a real-time environment.

### Tactile Texture Dataset – Data Collection
As described in (50), a spinning drum was designed to apply each texture to the sensor at a set scanning speed and applied force. Scanning speed was tested from 40 mm/s to 120 mm/s (in 20 mm/s increments) based on biological scanning speed measurements (51). These speeds correspond to trials of length 6, 4, 3, 2.4 and 2 s (from 40 mm/s to 120 mm/s). Applied force was measured through a load cell at the base of the drum and was tested at 250, 500, and 1000 g. The set of 5 speeds and 3 forces corresponds to 15 speed-force combinations. All 15 combinations were recorded for all 16 textures for 100 trials, making a dataset with 24,000 total trials. Each taxel was sampled with 10-bit resolution at 1 kHz. The taxel readings were normalized between 0 and 1 on a per-taxel basis (i.e. minimum and maximum readings for each taxel from the whole dataset was used). Control of the spinning drum was implemented on an Arduino Mega microcontroller (Arduino, Somerville, MA). Data recording and transmission was implemented on an Arduino Uno microcontroller (Arduino, Somerville, MA). More details about the system design, control, and validation can be found in (50).

### Tactile Texture Dataset – Force Invariance Module
All processing of data for the tactile texture dataset from this point was done in MATLAB R2023A (MathWorks, Natick, MA).

In order to create force invariant spiking patterns (25), the input current to the spiking activity encoding module is scaled to account for the applied force. There are a couple of complicating factors for the force scaling process. Firstly, the force reading is measured at a load cell at the base of the texture drum, so the actual force at the interface between the texture and sensor is unknown. Therefore, the 250, 500, and 1000 g force levels should be considered categories rather than actual force descriptors. Secondly, the relationship between applied force and sensor output is nonlinear because of saturation effects of the individual taxels and mechanical interaction effects due to the whole sensor array interfacing simultaneously with the drum. For these reasons, an empirical approach was used to determine the force scaling factors.

Since the goal is to create force invariant spiking patterns, the spike rate from the SA encoding of each taxel was used to make the objective function for an optimization problem. The target spike rate was set to the spike rate at the 500 g applied force (averaged across all trials). Therefore, scaling coefficients are determined for each texture for each taxel for each speed at 250 and 1000 g. The SA encoding of analog input to spiking activity is nonlinear but is monotonically increasing (i.e. higher input values produce more spikes), and therefore binary search is appropriate for finding the scaling coefficients. We solve for the matrix $C \in R^{16 \times 18 \times 5 \times 3}$ using equation (1). ε is set to 0.1 (i.e. 1 spike per 10 seconds). Using this method, only 62 out of 4320 coefficients (≈1.4%) did not converge and these corresponded to very low activity spike trains (≤ 2 spikes per second) where there are large discontinuities in the spiking activity encoding. For these cases, to avoid arbitrarily large scaling factors that would break the encoding module, the coefficients were limited to a value of 5.

**Tactile Texture Dataset – Spiking Activity Encoding Module**
The analog tactile sensor readings are transformed into neuromorphic spiking activity using the Izhikevich neuron model equations from (48) which are shown in equations (3), (4), and (5):

$$\frac{dv}{dt} = 0.04v^2 + 5v + 140 - u + kI$$

(3)

$$\frac{du}{dt} = a(bv - u)$$

(4)

$$\text{if } v \geq 30 \text{ mV, then } \begin{cases} v \leftarrow c \\ u \leftarrow u + d \end{cases}$$

(5)

These equations can model diverse behaviors of neuronal membrane voltage (v) found in biological neurons. Different behaviors are implemented by the choice of free parameters a, b, c, and d (52). The equations take a time-series current (I) and scaling factor (k) as input. When a neuron spikes (v ≥ 30 mV), the neuron's membrane voltage and recovery variable (u) are reset. In this work, we implement the SA mechanoreceptors using the Tonic Spiking model (a = 0.02, b = 0.2, c = -65, and d = 6) applied to the tactile sensor readings (scaled by $C_{i,j,k}^T$). The RA mechanoreceptors are implemented using the Tonic Spiking model as well but applied to the tactile sensor readings that have been smoothed with a 20 Hz lowpass filter and then differentiated. The scaling factor k is 100 and 3 for the SA and RA models, respectively. The analog output of the Izhikevich model is turned into a spike train vector which is an ordered list of spike times (i.e. times when v ≥ 30 mV). With 18 taxels (3x3x2) and 2 neurons models, 36 spike trains are generated per trial.

**Tactile Texture Dataset – Feature Extraction**
For spike trains produced by the SA encoding, the spike rate (SR) is calculated over non-overlapping sliding windows of length 100 ms. For spike trains produced by the RA encoding, the spike count (SC) is calculated over non-overlapping sliding windows of length 100 ms. For speed scaled data, all trials are 2 seconds long (the length of trials for the reference speed 120 mm/s), therefore each spike train produces 20 features. Since more spikes are produced by the SA encoding for longer trials, the SR must be normalized by the speed scaling coefficient as well to make the SR of the windows match between different trial speeds. For data that is not speed scaled, the number of features available depends on the trial length (which varies from 2 to 6 seconds). For classification purposes, all the trial feature lengths must be the same, so 60 features are output with zeros padding the shorter trials. With 36 spike trains produced per trial (18 taxels by 2 encodings), the total feature vector length is 720 and 2160 for speed scaled and non-speed scaled data, respectively. With 100 trials for 16 textures and 15 speed-force conditions, the total feature matrix dataset has size 24,000 x 720 or 24,000 x 2160 for speed scaled and non-speed scaled data, respectively.

**Tactile Texture Dataset – Statistical Analysis**
All classification is done using linear discriminant analysis (LDA) (54) on the set of PCs calculated from the feature data. To save on computation time, 500 trials are randomly selected for each texture (out of 1500) to make a feature matrix composed of 8000 trials. Additionally, to ensure robust results, k-fold cross-validation is done with k=4 (55) (i.e. 75%/25% training/testing split). Therefore, for each fold, the training set has 6000 trials, and the testing set has 2000 trials.

For the texture extrapolation classification task, the training and testing process is different because the training set is a subset of the speed-force combinations. Only 3 speeds (40, 60 and 80 mm/s) and 2 forces (250 and 500 g) are used for training which corresponds to 6 of the 15 speed-force combinations. The test set comes from all 15 speed-force combinations. Since the training set and test set are explicitly sampled from different distributions, k-fold cross-validation is not used in this case. For the trained speed-force combinations, 75 of the trials (per texture) are randomly chosen for training and the remaining 25 trials are used for testing. The training set for this task has 7200 trials (16 textures x 75 trials x 6 speed-force combinations). For the speed-force combinations only used for testing, 25 trials are randomly chosen for testing.

For all tasks, for each number of PCs, the classification is done 20 times to get mean (µ) and standard deviation (σ) statistics for accuracy. Therefore, all mean and standard deviation results in Fig. 3 and Table 1 are for n = 20. The underlying tested trials in Fig. 3A-B are 160,000 (8000 trials x 20 iterations). The underlying tested trials in Fig. 3C-F are [16,000 24,000 32,000 48,000], respectively (16 textures x 25 trials x 20 iterations x [2 3 4 6] speed-force combinations).

**Real-time Texture Classification – Processing Pipeline and Training Data**
Flat rigid texture plates were 3D-printed corresponding to textures A (smooth), B (circular ridges), D (circular ridges with double space), L (waves), and N (waves with double space) in Fig. 2B. Each texture plate was 300 mm by 36 mm with the middle 200 mm of the texture plate being used for the classification experiment. A human operator applied a 3x3 tactile sensor array attached to a 3D-printed finger across the textures with an infrared motion tracker (HTC Vive, Taoyuan City, Taiwan) attached to their wrist and wirelessly connected to a computer running MATLAB R2023A (MathWorks, Natick, MA). On the computer, the average velocity was calculated every 100 ms from the motion tracker's position data and sent to a Teensy 4.0 microcontroller (PJRC, Sherwood, OR). The microcontroller also receives a stream of tactile data from the sensor array and runs the spiking activity encoding module and speed invariance module in real-time. In this case, the spiking activity encoding module only includes the SA model. Spike rate features are extracted from the spike trains and are sent to the computer to be classified once the finger has reached the end of the texture plate (determined by the position information from the motion tracker). For each trial, the operator was asked to move the finger across the texture plate in one of 5 general velocity profiles: slow, medium, fast, slow-to-fast, and fast-to-slow.

An initial dataset of 10 trials per texture was collected to get the minimum and maximum values for each taxel. These values were used to normalize the sensor readings between 0 and 1 (on a per-taxel basis) before the spiking activity encoding module. A training dataset of 20 trials per texture (4 trials per velocity profile) was collected to train the LDA model used for real-time classification.

**Real-time Texture Classification – Speed Invariance Module**
The speed invariance module in the real-time experiment operates similarly to the tactile texture dataset but has a couple key differences. First, the reference speed used in the real-time experiment is 100 mm/s (compared to 120 mm/s before). This is an arbitrary change (it mostly affects the spatial resolution of your extracted features) but was chosen to match the update rate of the velocity from the motion tracker. The more crucial difference is that in the real-time experiment, the velocity is changing dynamically, so the scaling of spike times must be done incrementally in batches (whereas the tactile texture dataset had a constant velocity so everything could be scaled with the same value). To do this, the speed scaling is computed after 100 ms of "real-time" spiking data is collected. The time gaps between spikes are scaled using equation (2) and the computed velocity of that window. With 9 taxels (3x3) and one neuron model, 9 "scaled-time" spike trains are generated per trial.

**Real-time Texture Classification – Feature Extraction**
The spike rate (SR) is calculated over non-overlapping sliding windows of length 100 ms (in "scaled-time"). For speed scaled data, the "scaled-time" length of a trial is 2 seconds and therefore each trial produces 20 features. The speed scaled SR has to be normalized by a coefficient representing the "real-time" duration of that "scaled-time" window to make sure the SRs match between different velocity profiles. For data that is not speed scaled, the number of features depends on the trial length, so the feature vector is either cut down to the last 20 features or padded with zeros to reach 20 features. With 9 spike trains produced per trial, the total feature vector length is 180. With 20 training trials for 5 textures, the total training feature matrix dataset is 100 x 180. The feature matrix is mean-centered to reduce the effect of session-to-session variability in the sensor data.

**Real-time Texture Classification – Protocol**
The real-time texture classification protocol consists of several steps. First, an LDA model was trained using 25 PCs computed from the mean-centered feature matrix from the training dataset. This number of PCs was selected because, in the offline dataset, the classification accuracies typically had reached their plateau. Since there is session-to-session variability in the sensor data, the mean vector (subtracted from the feature matrix to center it) is calibrated right before real-time classification. To do this, the mean vector is calculated from a new set of 10 trials for each texture (each velocity profile is performed twice). This mean vector is used to mean-center the real-time feature data before calculating the first 25 PCs. The real-time dataset consists of 20 trials for each texture (each velocity profile is performed 4 times) which is then classified with the trained LDA model.

**Real-time Texture Classification – Statistical Analysis**
For the offline classification results (Fig. 4B blue lines), an LDA model was trained on a set of PCs calculated from feature data (composed of 100 trials). To ensure robust results, k-fold cross-validation is done with k=4 (55) (i.e. 75%/25% training/testing split). Therefore, for each fold, the training set has 75 trials, and the testing set has 25 trials.

For the real-time classification results (Fig. 4B green lines), 75% of trials chosen randomly from a previously collected training dataset are used to train the LDA model. Since the training set and test set (collected and tested in real-time) are explicitly sampled from different distributions, k-fold cross-validation is not used in this case. Therefore, the training set has 75 trials, and the testing set has 100 trials.

For the extrapolation classification results (Fig. 4C), the LDA model is trained with 75% of trials randomly chosen from a subset of the velocity profiles in the previously collected training dataset. In this case, 2 velocity profiles (slow, medium) are used for training, corresponding to 30 trials (5 textures x 4 trials x 2 velocities x 75%). Since the training set and test set (collected and tested in real-time) are explicitly sampled from different distributions, k-fold cross-validation is not used in this case. In the test set, the trained velocity profiles (blue lines) correspond to 40 trials. The untrained velocity profiles (fast, slow-to-fast, fast-to-slow) correspond to 60 trials (green lines).

For each of these tasks, 3 datasets were collected and tested. For each dataset and number of PCs, the classification is done 100 times to get mean ($\mu$) and standard error ($\sigma_{\bar{x}}$) statistics for accuracy. All mean and standard error results in Fig. 4B-C are for n = 300. The underlying tested trials in Fig. 4B are 30,000 (100 iterations x 3 datasets x 100 trials). The underlying tested trials in Fig. 4C are [12,000 18,000] for blue and green lines, respectively (100 iterations x 3 datasets x [40 60] trials).

Significance between the "Original" data and "Speed Scaled" data are calculated for each task at each PC using the two-tailed, two-population Welch's t-test (56). Cohen's d is used to calculate effect size (57). A single asterisk corresponds to $p < 0.05$ and a "medium" effect size ($d > 0.5$). Double asterisks correspond to $p < 0.01$ and a "large" effect size ($d > 1$).

Finally, the real-time classifier is deployed for the demonstration shown in Movie S3. To maximize accuracy, the LDA model is trained with all 100 trials of the training dataset. The real-time texture classification demonstration was repeated 3 times each with and without the speed invariance module. The demonstration was done with 25 PCs, but the experiment was simulated offline for all other number of PCs. The mean accuracies in Fig. S5 are for n=300 (100 trials x 3 datasets). Significance between the "Original" data and "Speed Scaled" data are calculated at each PC using a two-tailed, two-proportion z-test (58). Cohen's h is used to calculate effect size (59). A single asterisk corresponds to $p < 0.05$ and a "small" effect size ($h > 0.2$). Double asterisks correspond to $p < 0.01$ and a "medium" effect size ($d > 0.5$).

## Acknowledgments

We thank K. Ding, W. Mbuguiro, L. E. Osborn, K. Quinn, and M. Rakhshan for their comments on the manuscript.

**Figures and Tables**

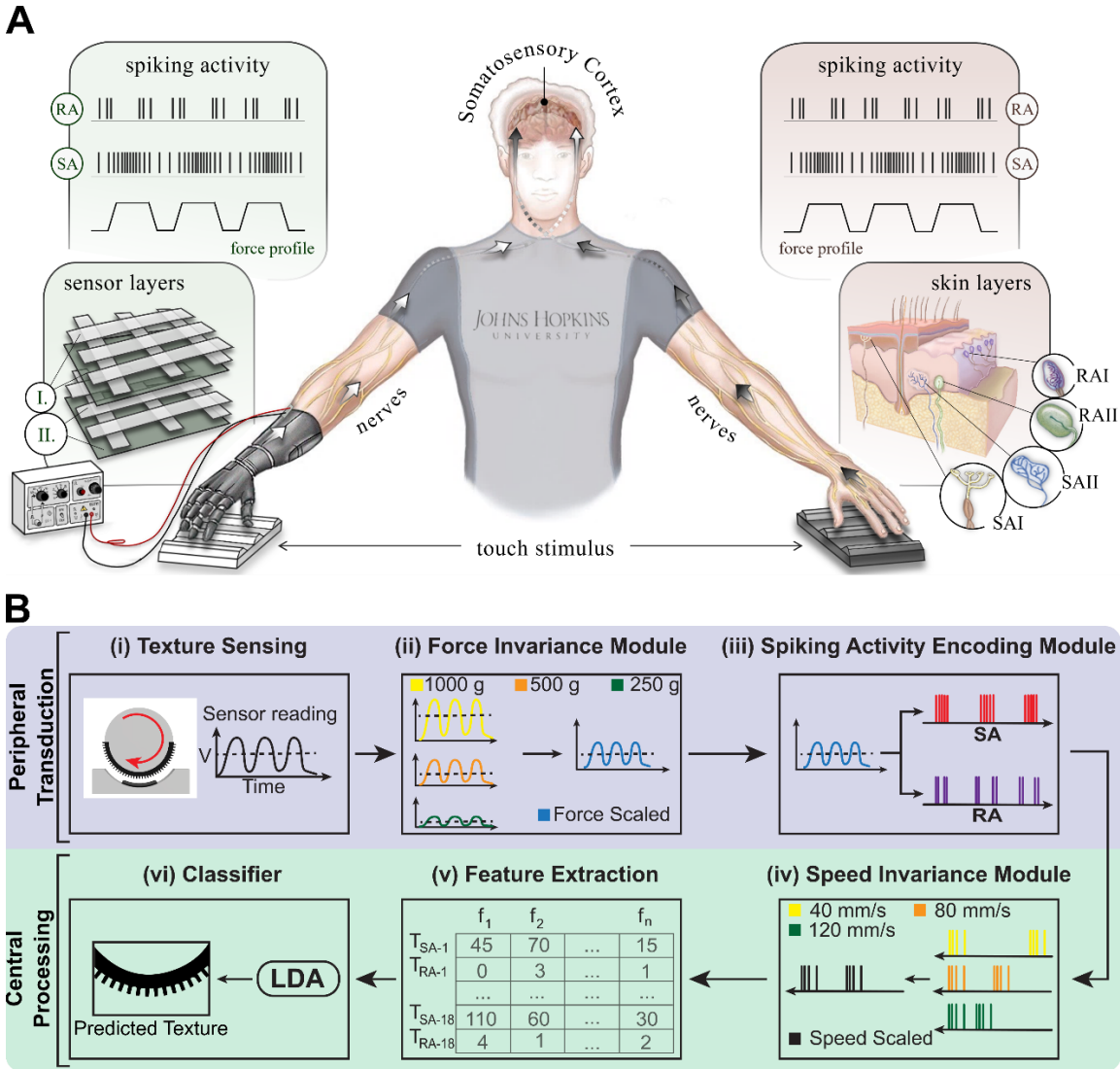

**Figure 1. Potential Neuromorphic Tactile Sensory Feedback in a Neuroprosthesis and the presently implemented signal processing pipeline. (A)** The overall goal of the artificial sensory feedback system (left) is to restore naturalistic touch sensation by mimicking biological tactile processing (right). The artificial and biological pathways start with transduction of tactile stimuli into neuronal spiking activity (piezoresistive tactile sensor and mechanoreceptors in skin, respectively). Biological skin has SAI, SAII, RAI, and RAII mechanoreceptors. The tactile sensor has two layers corresponding to type I and type II mechanoreceptors which are then encoded with SA and RA models to mimic all four types of mechanoreceptors. A stimulator (bottom left) electrically activates the nerve according to the neuromorphic encoding (44) to produce scanning speed and contact force invariant representations of texture in the somatosensory cortex which are perceived as natural by the prosthesis user. **(B)** The signal processing stages of the neuromorphic encoding process used in this work. Stages (i) - (iii) correspond to the transduction of tactile stimuli into force-invariant spiking activity (as in peripheral transduction). Stages (iv) - (vi) correspond to the classification of texture after augmenting with speed-invariance (as in biological central processing and perception). (i) The rotating texture drum pressed against a tactile sensor

produces analog sensor readings. (ii) Analog readings are scaled to create force-invariant signals. (iii) The analog signals are then transformed to SA and RA spiking activity using the Izhikevich Neuron Model (48, 52). (iv) The spiking activity is scaled in time to produce speed and force-invariant spiking patterns. (v) Spiking features ($f_1$ to $f_n$) are calculated for each taxel and spike encoding ($T_{SA-1}$ to $T_{RA-18}$) over a series of non-overlapping sliding time windows. (vi) The extracted features are passed through a trained linear discriminant analysis (LDA) model to predict the texture stimulus.

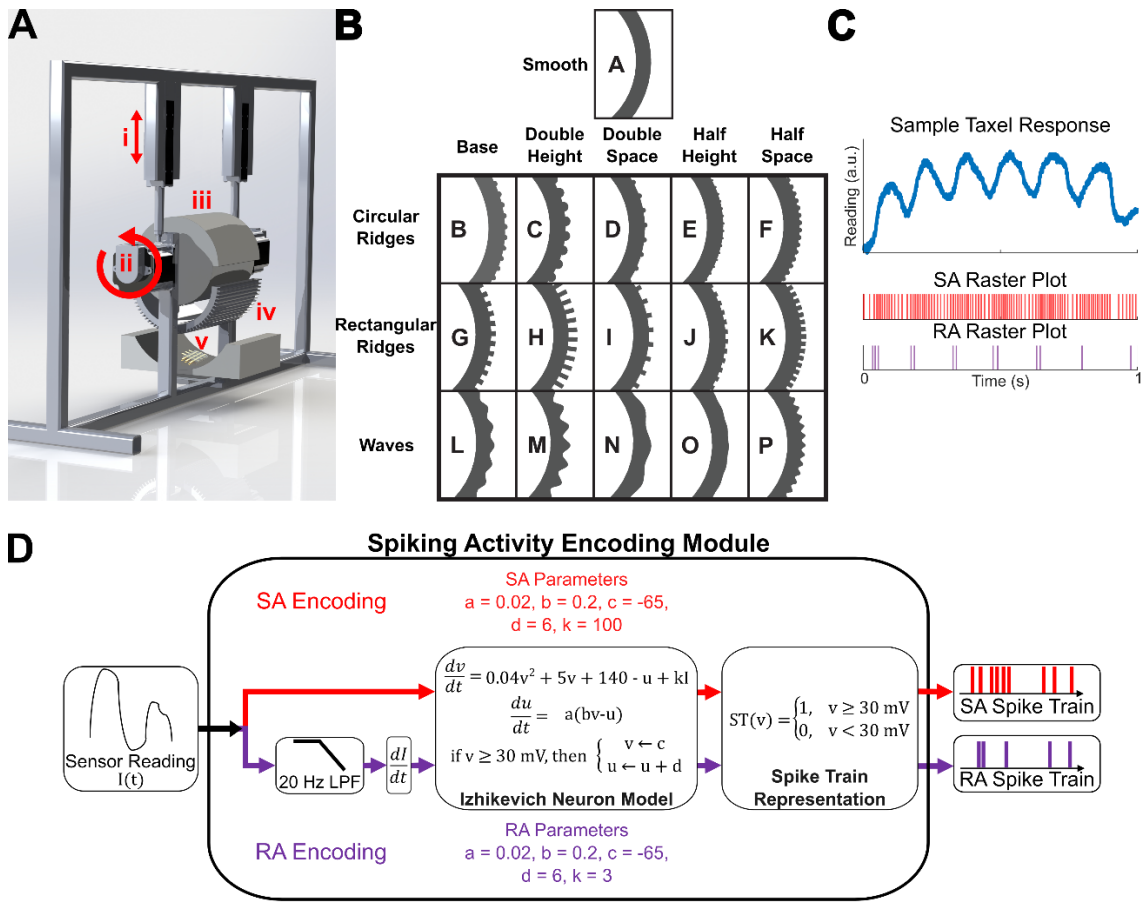

**Figure 2. Robotic Drum, Texture Set, and Neuromorphic Encoding. (A)** The tactile stimulator robot is used for data collection and can be programmed to maintain precise scanning speed and contact force. (i) An actuator moves the texture drum vertically to reach a desired force setpoint. (ii) A stepper motor rotates the drum at a programmed constant velocity. (iii) The drum applies a texture stimulus to a tactile sensor. (iv) An example 3D-printed texture plate that can be affixed to the drum. (v) 3x3x2 tactile sensor that is measuring the tactile stimulus. **(B)** The tactile dataset has 16 textures belonging to 3 texture groups (with 5 variations each) and a smooth control texture. The variations in textures are designed to affect the amplitude and period of tactile readings and therefore confound speed and force invariance algorithms. **(C)** An example of the encoding of analog readings from a taxel into SA and RA spiking activity using the spiking activity encoding module. The SA raster plot shows that the spike rate is proportional to the amplitude of the taxel signal. The RA raster plot shows that spikes are produced when there are large changes in the taxel signal. This example is 1 second of data from the base waves texture (texture L) from 1 taxel from 1 trial at 500 g applied force and 80 mm/s scanning speed. **(D)** The Spiking Activity Encoding Module takes an analog time series of sensor readings I(t) and outputs two spike trains representing the SA and RA encoding of the data. The SA encoding applies the Tonic Spiking model to the sensor readings to produce spiking activity at a rate proportional to sensor readings. The RA encoding first smooths (20 Hz LPF) and differentiates the data and then applies the Tonic Spiking model to produce spiking activity at a rate proportional to changes in the sensor readings. The Izhikevich model outputs an analog membrane voltage v(t) representing the membrane voltage of a neuron. A spike train is a binarized representation of the membrane voltage. A 'spike' (represented as 1) occurs when v(t)≥30 mV and is 0 otherwise. The spike train can be equivalently represented as a vector of spike times (the indices where a 'spike' occurs). In this work, a vector of spike times is used because it is a more compact representation of a spike train, and the calculation is simpler for the Speed Invariance Module.

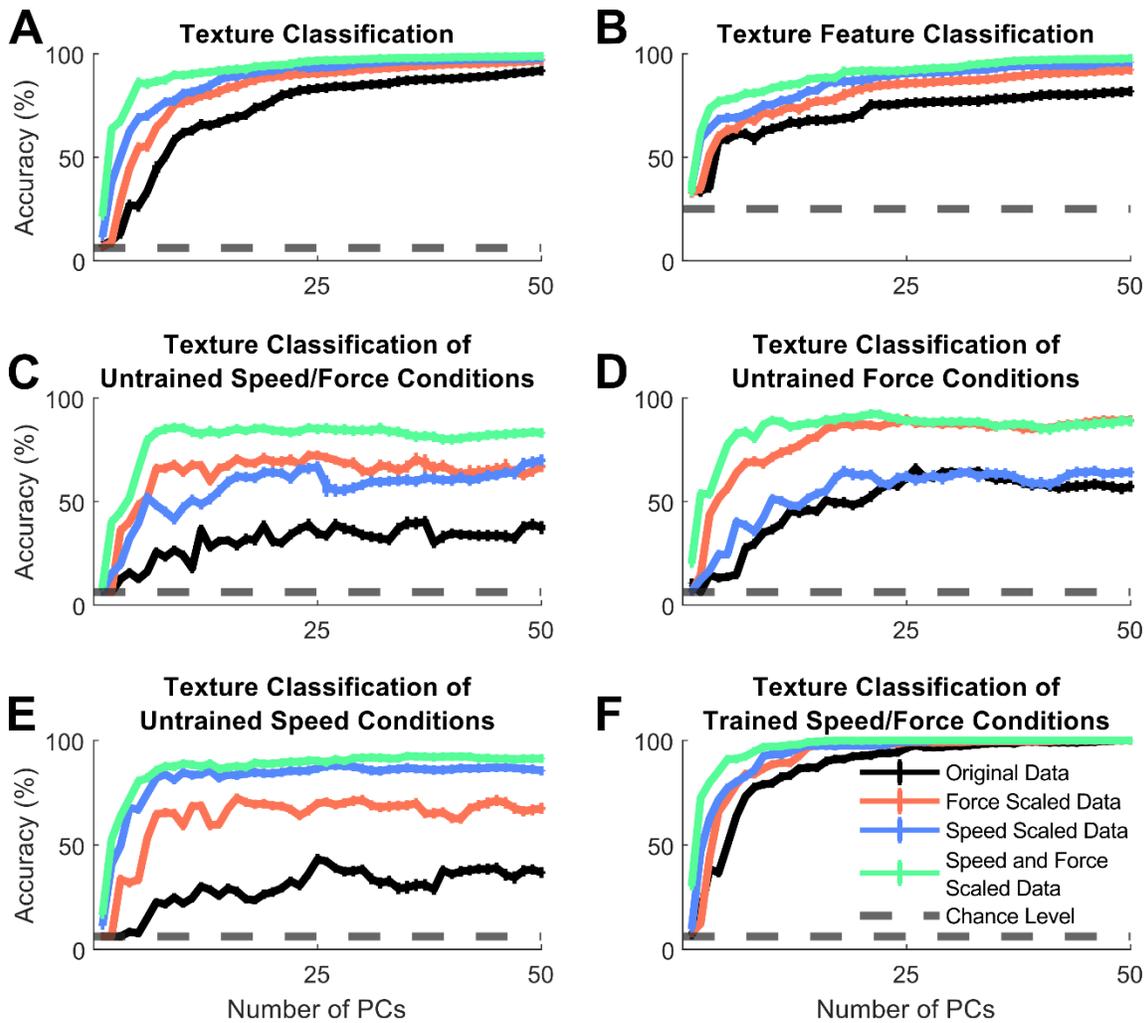

**Figure 3. Offline texture classification accuracies for different tasks.** For each task, classification accuracy is computed without the speed and force invariance modules included (black), with only force invariance included (red), with only speed invariance included (blue), and with both speed and force invariance included (green). The accuracy is computed over a range of principal components (from 1 to 50) used in the classifier. The datapoints of the line plots (n = 20) are the mean classification accuracies (μ), and vertical error bars represent the standard deviation (σ). In (A) and (B) the classifier is trained with all 15 speed-force conditions. **(A)** Classifier is asked to predict one of the 16 textures (chance level is 6.25%). **(B)** Classifier is asked to predict one of the 4 texture groups (chance level is 25%). In (C) – (F) the classifier is trained with a subset of the speed-force conditions and asked to predict one of the 16 textures (chance level is 6.25%). Three speeds were trained (40, 60, 80 mm/s) and two were untrained (100, 120 mm/s). Two forces were trained (250, 500 g) and one was untrained (1000 g). **(C)** Classification of trials where both speed and force conditions were not in the training set. **(D)** Classification of trials where speed condition was in the training set, but force condition was not. **(E)** Classification of trials where force condition was in the training set, but speed condition was not. **(F)** Classification of trials where both speed and force conditions were in the training set.

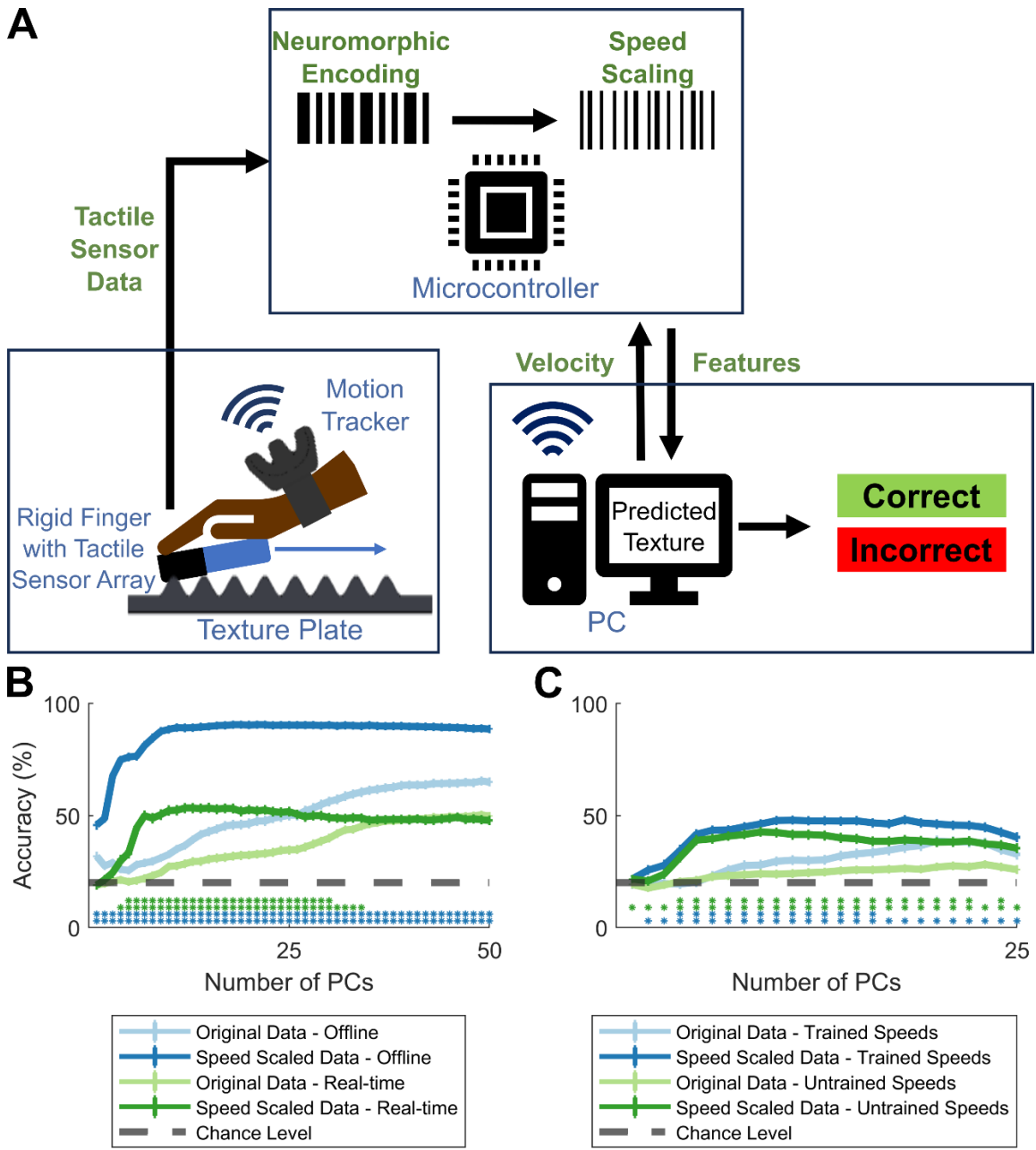

**Figure 4. Real-time Texture Classification System and Analysis. (A)** An operator wearing a motion tracker moves a 3x3 tactile sensor attached to a rigid finger across the texture plate. The tactile data goes to a microcontroller to be processed. The position data is sent wirelessly to a computer. A microcontroller uses the tactile and velocity data streams for the spiking activity encoding and speed invariance modules. The spike train features are extracted and sent to the computer for classification. The computer calculates the velocity of the sensor from the motion tracker and sends it to the microcontroller. It also receives spike train features from the microcontroller and classifies the texture in real-time. The predicted texture is displayed on the computer monitor. The background color indicates whether the predicted texture matches the ground truth. In (B) and (C), a classifier is asked to predict one of 5 textures (chance level is 20%). For each task, classification accuracy is computed with and without the speed invariance module included (darker and lighter lines, respectively). The accuracy is computed over a range

of principal components used in the classifier. The datapoints of the line plots (n = 300) are the mean classification accuracies (μ), and vertical error bars represent the standard error ($\sigma_{\bar{x}}$). Significance and effect size are computed between the "Original" and "Speed Scaled" datasets (56, 57). A single asterisk corresponds to p < 0.05 and a "medium" effect size (d > 0.5). Double asterisks correspond to p < 0.01 and a "large" effect size (d > 1). **(B)** An initial offline dataset (blue lines) is collected to train the classifier. This model is then used to classify texture data from a new session (green lines). **(C)** The classifier is trained with a subset of the velocity profiles in the training dataset. Two speeds were trained (slow, medium) and three were untrained (fast, slow-to-fast, fast-to-slow). The blue lines correspond to trials in a new test set in the trained velocity profiles. The green lines correspond to trials in the test set in the untrained velocity profiles.

**Table 1. Classification Accuracy with 50 PCs (%).**

|  | Original | | Speed Scaled | | Force Scaled | | Speed and Force Scaled | |
|---|---|---|---|---|---|---|---|---|
|  | μ (%) | σ (%) | μ (%) | σ (%) | μ (%) | σ (%) | μ (%) | σ (%) |
| Individual Textures (Fig. 3A) | 91.78 | 0.55 | 97.76 | 0.14 | 96.84 | 0.43 | 98.85 | 0.16 |
| Texture Groups (Fig. 3B) | 81.83 | 0.53 | 95.89 | 0.26 | 92.20 | 0.43 | 97.78 | 0.19 |
| Untrained Speed and Force (Fig. 3C) | 37.44 | 1.28 | 69.98 | 0.87 | 67.01 | 0.85 | 83.15 | 0.80 |
| Untrained Force (Fig. 3D) | 57.35 | 0.79 | 64.18 | 0.55 | 89.38 | 0.43 | 88.73 | 0.52 |
| Untrained Speed (Fig. 3E) | 36.81 | 0.66 | 85.52 | 0.34 | 67.53 | 0.78 | 91.50 | 0.27 |
| Trained Speed and Force (Fig. 3F) | 99.93 | 0.04 | 99.95 | 0.04 | 99.91 | 0.06 | 99.96 | 0.04 |

# Supporting Information for
Invariant neuromorphic representations of tactile stimuli improve robustness of a real-time texture classification system

Mark M. Iskarous[*], Zan Chaudhry, Fangjie Li, Samuel Bello, Sriramana Sankar, Ariel Slepyan, Natasha Chugh, Christopher L. Hunt, Rebecca J. Greene, Nitish V. Thakor

Corresponding Author: Mark M. Iskarous
Email: miskarous@uchicago.edu

**This PDF file includes:**

    Figures S1 to S5
    Legends for Movies S1 to S3
    SI References

**Other supporting materials for this manuscript include the following:**

    Movies S1 to S3

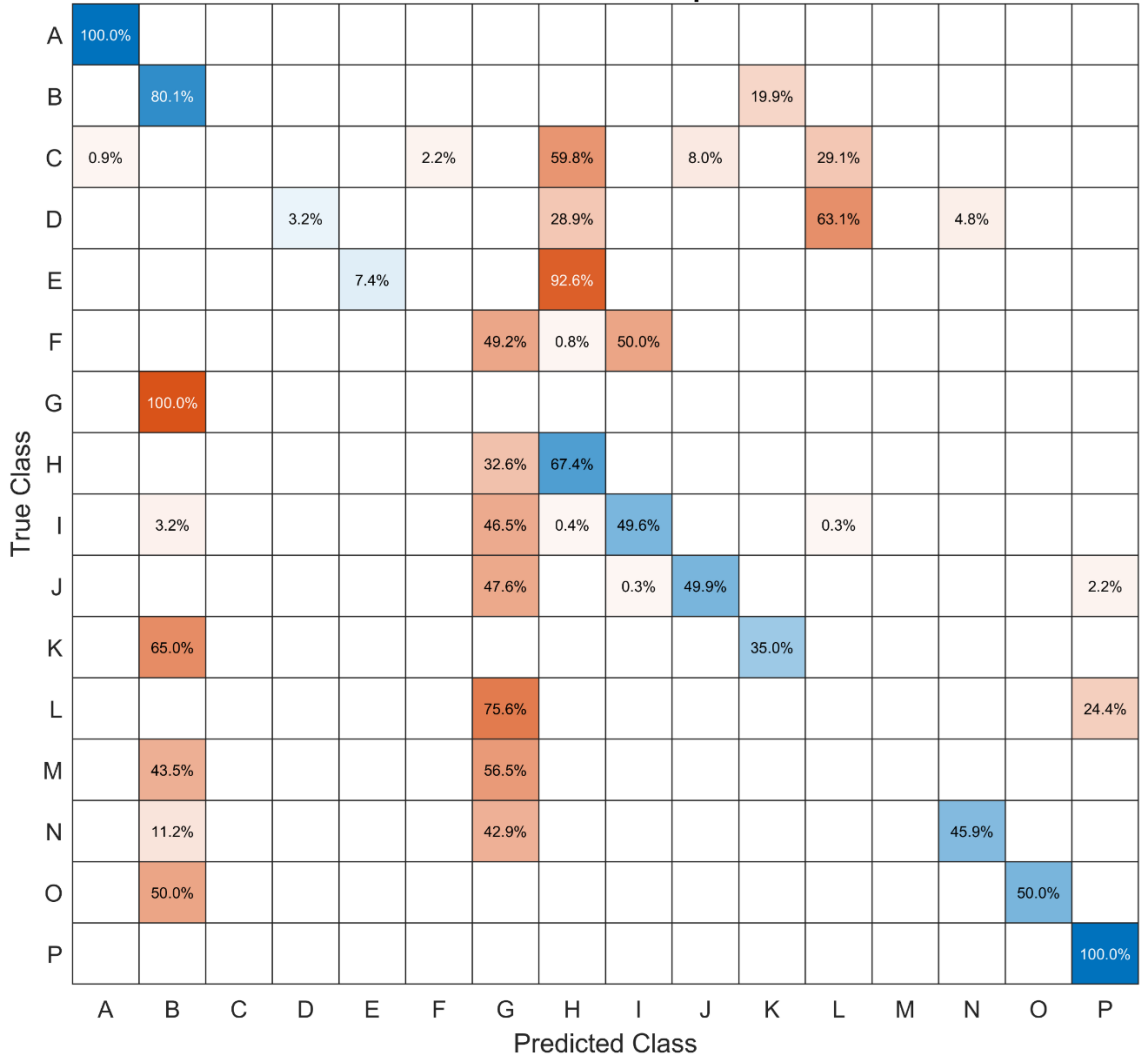

**Fig. S1. Confusion Matrix for Texture Extrapolation Classification on "Original" Data for Untrained Speed/Force Conditions.** This confusion matrix breaks down the results from the texture extrapolation classification analysis at 50 PCs when confronted simultaneously with novel speed and force conditions (Fig. 3C). Specifically, this classifier was trained using the "Original" data. The class labels match Fig. 2B. Since there are two novel speeds (100 and 120 mm/s) and one novel force (1000 g) there are two novel speed-force combinations tested. Therefore, each texture is tested 1000 times (25 trials x 20 iterations x 2 speed-force combinations). Overall classification accuracy for this set is 36.78% (for 16,000 total test trials).

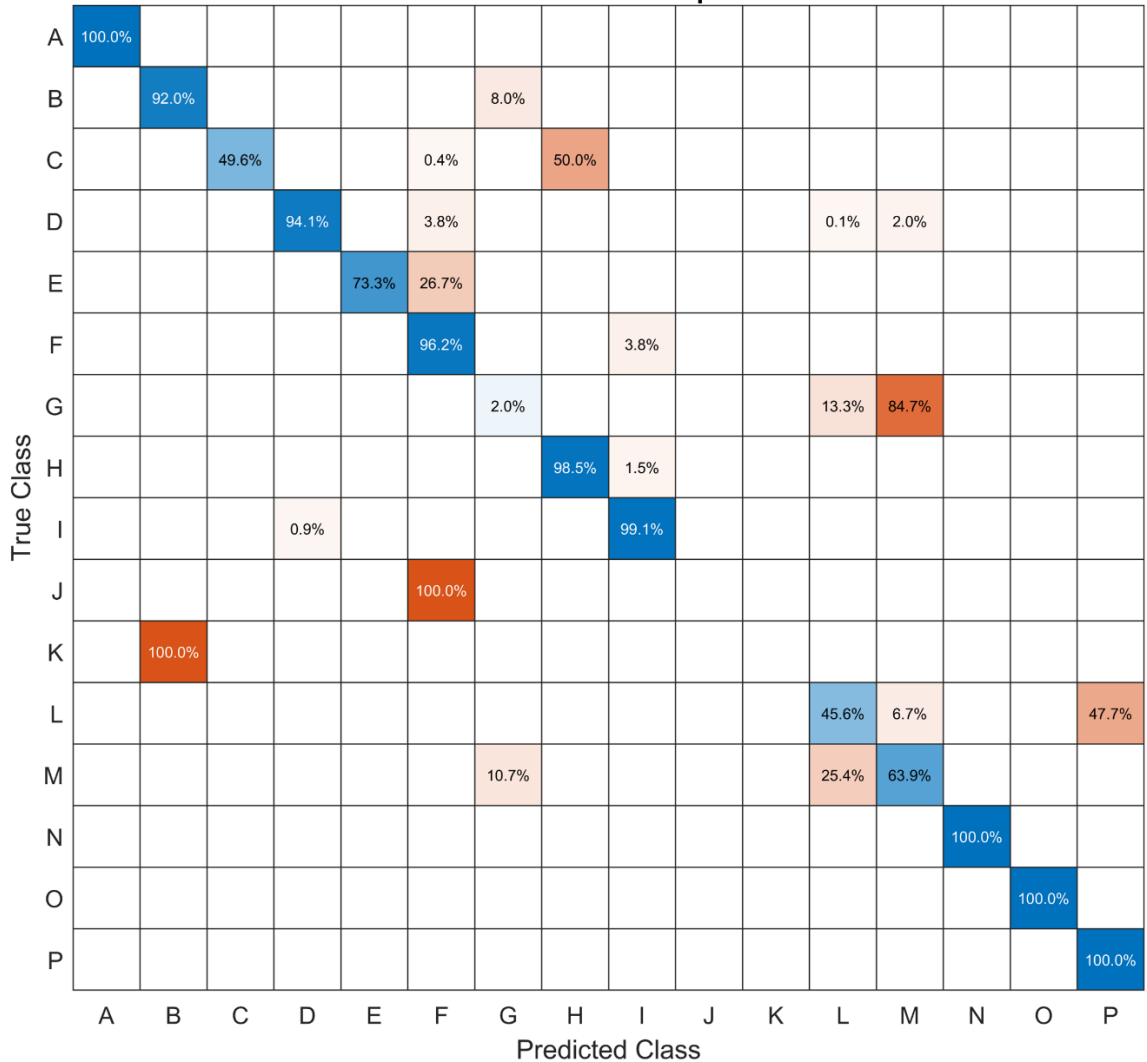

**Fig. S2. Confusion Matrix for Texture Extrapolation Classification on "Speed Scaled" Data for Untrained Speed/Force Conditions.** This confusion matrix breaks down the results from the texture extrapolation classification analysis at 50 PCs when confronted simultaneously with novel speed and force conditions (Fig. 3C). Specifically, this classifier was trained using the "Speed Scaled" data. The class labels match Fig. 2B. Since there are two novel speeds (100 and 120 mm/s) and one novel force (1000 g) there are two novel speed-force combinations tested. Therefore, each texture is tested 1000 times (25 trials x 20 iterations x 2 speed-force combinations). Overall classification accuracy for this set is 69.64% (for 16,000 total test trials).

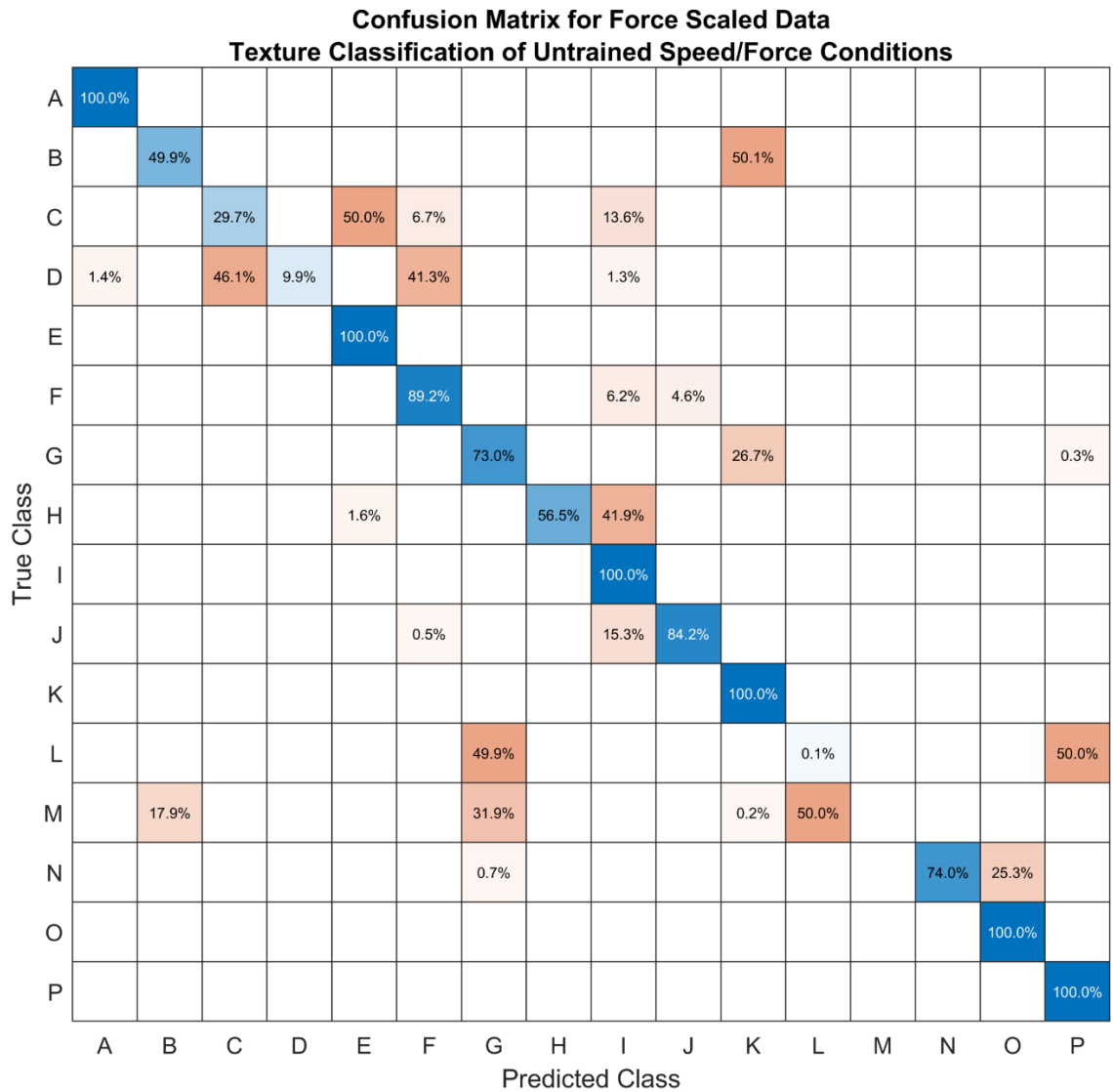

**Fig. S3. Confusion Matrix for Texture Extrapolation Classification on "Force Scaled" Data for Untrained Speed/Force Conditions.** This confusion matrix breaks down the results from the texture extrapolation classification analysis at 50 PCs when confronted simultaneously with novel speed and force conditions (Fig. 3C). Specifically, this classifier was trained using the "Force Scaled" data. The class labels match Fig. 2B. Since there are two novel speeds (100 and 120 mm/s) and one novel force (1000 g) there are two novel speed-force combinations tested. Therefore, each texture is tested 1000 times (25 trials x 20 iterations x 2 speed-force combinations). Overall classification accuracy for this set is 66.66% (for 16,000 total test trials).

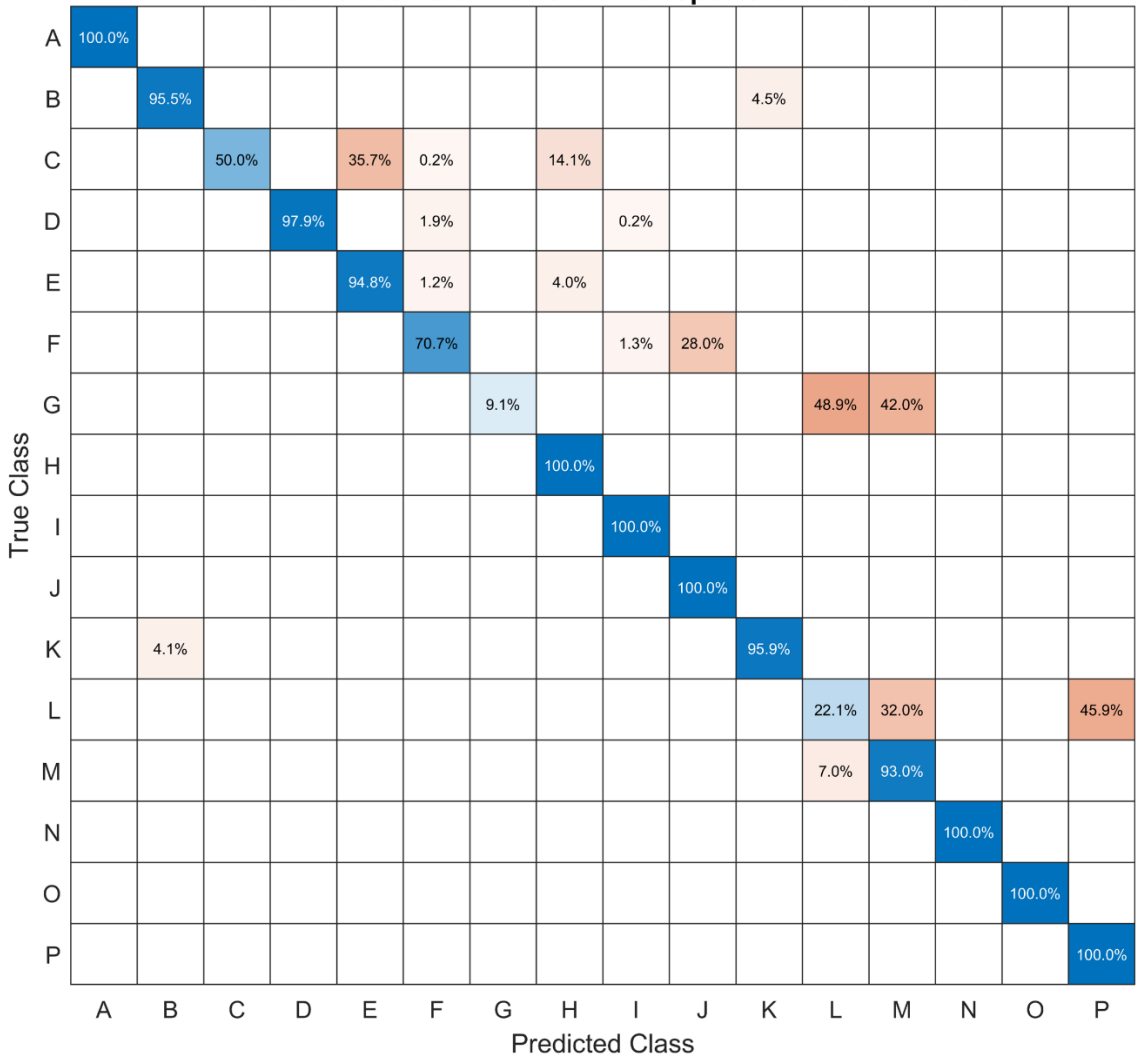

**Fig. S4. Confusion Matrix for Texture Extrapolation Classification on "Speed and Force Scaled" Data for Untrained Speed/Force Conditions.** This confusion matrix breaks down the results from the texture extrapolation classification analysis at 50 PCs when confronted simultaneously with novel speed and force conditions (Fig. 3C). Specifically, this classifier was trained using the "Speed and Force Scaled" data. The class labels match Fig. 2B. Since there are two novel speeds (100 and 120 mm/s) and one novel force (1000 g) there are two novel speed-force combinations tested. Therefore, each texture is tested 1000 times (25 trials x 20 iterations x 2 speed-force combinations). Overall classification accuracy for this set is 83.06% (for 16,000 total test trials).

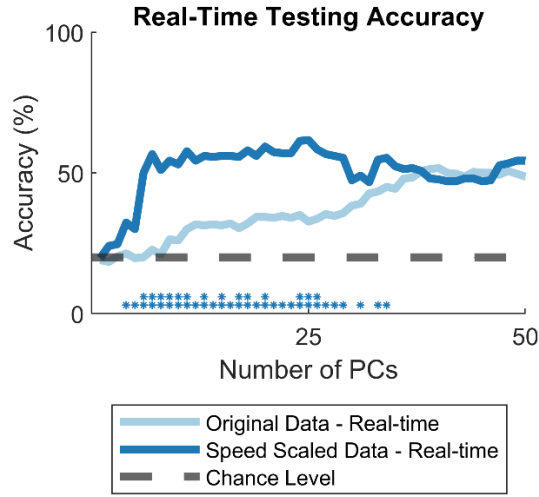

**Fig. S5. Real-time Texture Classification Demo Results.** The real-time system uses a classifier to predict one of five textures (chance level is 20%). The classifier was trained on all 100 trials of a training set that has five velocity profiles. Classification accuracy is computed with and without the speed invariance module included (darker and lighter lines, respectively). The accuracy is computed over a range of principal components used in the classifier. The datapoints of the line plots (n = 300) are the mean classification accuracies (µ). Significance and effect size are computed between the "Original" and "Speed Scaled" datasets (1, 2). A single asterisk corresponds to $p < 0.05$ and a "small" effect size ($h > 0.2$). Double asterisks correspond to $p < 0.01$ and a "medium" effect size ($d > 0.5$).

**Movie S1 (separate file). Rotating Drum Apparatus Data Collection**

**Movie S2 (separate file). Example Neuromorphic Encoding**

**Movie S3 (separate file). Real-time Texture Classification with Speed Invariant Neuromorphic Representation**

**SI References**

1. "Ch. 23: Comparing Two Proportions" in *The Basic Practice of Statistics*, (Macmillan Higher Education, 2017), pp. 950–987.

2. "Ch. 6: Differences between Proportions" in *Statistical Power Analysis for the Behavioral Sciences*, 2nd Ed., (Routledge, 1988), pp. 179–214.